\documentclass{article}

    \usepackage[final]{bdl_2021_camera_ready}

\usepackage[utf8]{inputenc} 
\usepackage[T1]{fontenc}    
\usepackage{hyperref}       
\usepackage{url}            
\usepackage{booktabs}       
\usepackage{amsfonts}       
\usepackage{nicefrac}       
\usepackage{microtype}      
\usepackage{xcolor}         
\usepackage{amsmath}
\usepackage{amssymb}
\usepackage{hhline}
\usepackage{graphicx}
\usepackage{caption,subcaption}

\usepackage{multirow}

\makeatletter

\usepackage{tikz}
\usetikzlibrary{shapes.misc, positioning}
\usetikzlibrary{backgrounds,fit,matrix, calc}
\tikzset{
   box/.style = {minimum height=10pt, minimum width=10pt, draw, rounded corners,rectangle, fill=white!50},
}
\usetikzlibrary{fadings}

\tikzset{
   boxconv/.style = {minimum height=2cm, minimum width=2cm, draw, line width=0.4mm, fill opacity=0.9, rounded corners,rectangle, fill=white!50},
}

\tikzset{
   boxconv_inactive/.style = {minimum height=2cm, minimum width=2cm,line width=0.3mm, draw, line width=0.1mm , fill opacity=0.9, rounded corners,rectangle, gray, fill=white!50},
}
\tikzset{
   input/.style = {minimum height=3cm, minimum width=3cm, draw, , fill opacity=0.9, rectangle, fill=white!50},
}

\tikzset{
   boxpooled/.style = {minimum height=1.5cm, minimum width=1.5cm, draw, line width=0.4mm, fill opacity=0.9, rounded corners,rectangle, fill=white!50},
}

\tikzset{
   boxpooled_inactive/.style = {minimum height=1.5cm, minimum width=1.5cm, draw, line width=0.4mm, fill opacity=0.9, rounded corners,rectangle, gray, fill=white!50},
}

\usetikzlibrary{fadings}\tikzset{
    boxwta/.style={%
        draw=black, thick,
        rectangle,
        rounded corners,
        minimum height=3cm,
        minimum width=3cm
    }
}

\usetikzlibrary{fadings}\tikzset{
    box1/.style={%
        draw=black, thick,
        rectangle,
        minimum height=2cm,
        minimum width=2cm
    }
}

\usetikzlibrary{fadings}\tikzset{
    box2/.style={%
        draw=black, thick,
        rectangle,
        minimum height=1.cm,
        minimum width=1.cm
    }
}

\usetikzlibrary{fadings}\tikzset{
    box3/.style={%
        draw=black, thick,
        rectangle,
        minimum height=.8cm,
        minimum width=.8cm
    }
}
\usetikzlibrary{decorations.markings}

\usepackage{multirow}

\title{Stochastic Local Winner-Takes-All Networks Enable Profound Adversarial Robustness}

\author{%
	Konstantinos P.~Panousis \\
	Cyprus University of Technology\\
	Limassol, Cyprus \\
	\texttt{k.panousis@cut.ac.cy} \\
	\And
	Sotirios Chatzis\\
	Cyprus University of Technology\\
	Limassol, Cyprus \\
	\texttt{sotirios.chatzis@cut.ac.cy} \\
	\AND
	Sergios Theodoridis\\
	National and Kapodistrian University of Athens, Greece\\
	Aalborg University, Aalborg, Denmark\\
	\texttt{stheodor@di.uoa.gr} 
}
\hypersetup{draft}

\begin{document}
	
	\maketitle
	
	\begin{abstract}
		This work explores the potency of stochastic competition-based activations, namely Stochastic Local Winner-Takes-All (LWTA), against powerful (gradient-based) white-box and black-box adversarial attacks; we especially focus on Adversarial Training settings. In our work, we replace the conventional ReLU-based nonlinearities with blocks comprising locally and stochastically competing linear units. The output of each network layer now yields a sparse output, depending on the outcome of winner sampling in each block. We rely on the Variational Bayesian framework for training and inference; we incorporate conventional PGD-based adversarial training arguments to increase the overall adversarial robustness. As we experimentally show, the arising networks yield state-of-the-art robustness against powerful adversarial attacks while retaining very high classification rate in the benign case.
	\end{abstract}
	
	\section{Introduction}
	
	In this paper, we revisit the novel stochastic formulation of deep networks with LWTA activations  \citep{panousis2019nonparametric, panousis21a},  and delve deeper into its potency against adversarial attacks in the context of PGD-based Adversarial Training (AT) \citep{madry2017towards}. We evaluate the robustness of the emerging networks against powerful gradient-based white-box as well as black-box adversarial attacks using the well-known AutoAttack (AA) framework \citep{croce2020reliable}. We provide the related source code at: \url{https://github.com/konpanousis/Adversarial-LWTA-AutoAttack}. 
	We experimentally show that Stochastic LWTA-based networks not only yield state-of-the-art accuracy against all the considered attacks, but do so while retaining very high classification accuracy in the benign case.
	
	\section{Stochastic Local Winner-Takes-All (LWTA) Networks}

	Let us consider an input $\boldsymbol x\in \mathbb{R}^{J}$ presented to a conventional deep neural network layer comprising $K$ hidden units and weight matrix $\boldsymbol W \in \mathbb{R}^{J \times K}$. Each hidden unit $k$ within the layer performs an inner product computation $h_k=\boldsymbol w_k^T \boldsymbol x =\sum_{j=1}^J w_{jk} \cdot x_j \in \mathbb{R}$; this (usually) passes through a  non-linear function $\sigma(\cdot)$. Thus, the final layer output vector $\boldsymbol y \in \mathbb{R}^K$ is formed via the concatenation of the non-linear activation of each individual hidden unit, such that  $ \boldsymbol y = [y_1,\dots, y_K]$, where $y_k = \sigma(h_k)$. 
	
	In contrast, in a fully connected LWTA-based network layer, singular nonlinear units are replaced by $U$ linear competing units aggregated together in a (LWTA) block; in the following, we denote with $B$ the number of LWTA blocks in an LWTA-based layer. The associated weights are now arranged as a three dimensional matrix $\boldsymbol W \in \mathbb{R}^{J \times B \times U}$ signifying that the input $\boldsymbol x$ is presented to each block and each unit therein. Specifically, in the LWTA-based framework, the $u$\textsuperscript{th} \textit{competing} unit within the $b$\textsuperscript{th} block computes its activation $h_{b,u}$ via the standard inner product computation $h_{b,u} = \boldsymbol w_{b,u}^T \boldsymbol x= \sum_{j=1}^J w_{j,b,u} \cdot x_j \in\mathbb{R}$; then, competition takes place among the units in the block. 
	
	The underlying principle is that out of the $U$ units in an LWTA block, \textit{only one} can be the \textit{winner}; this unit gets to pass its \textit{linear activation} to the next layer, while the rest output zero values. Thus, the output of an LWTA layer $\boldsymbol y \in \mathbb{R}^{B\cdot U}$ is composed of $B$ subvectors $\boldsymbol y_b \in \mathbb{R}^U$, one for each LWTA block and each with a single non-zero entry. It is apparent that this competition process results in a \textit{sparse representation}, since all units, \textit{except one} in each block, produce a zero output. In related literature, the competition procedure is deterministic, i.e., the winner unit is the one with the highest activation. However, stochastic competition principles have been recently proposed in \cite{panousis2019nonparametric,panousis21a, Voskou_2021_ICCV}.
	\begin{figure}
		\centering
		\includegraphics[scale=1.1]{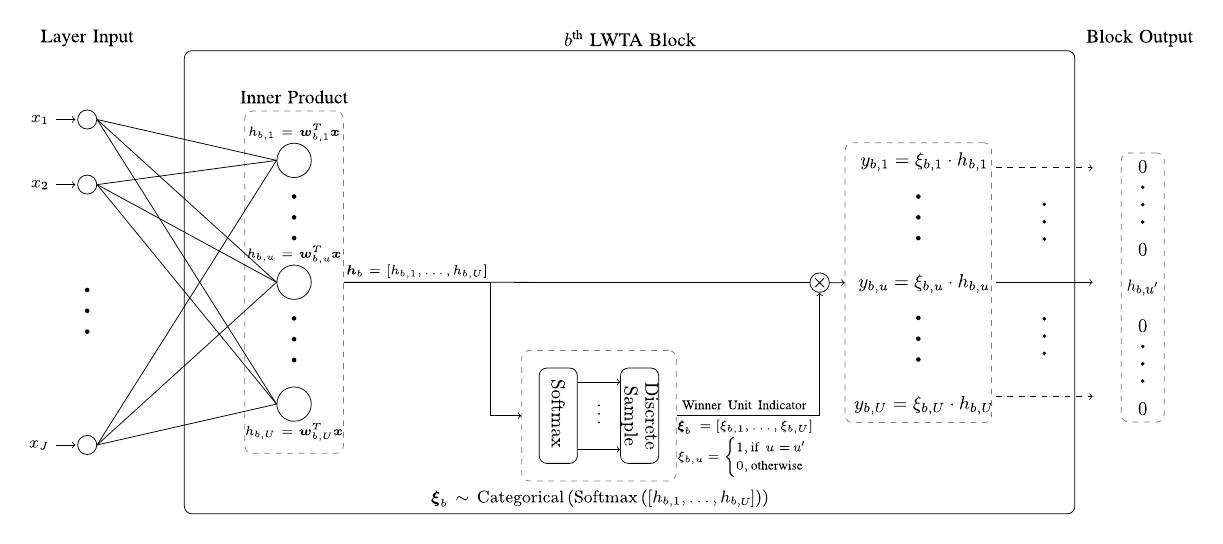}
		\caption{A detailed bisection of the $b$\textsuperscript{th} Stochastic LWTA block in an LWTA layer. Presented with an input $\boldsymbol x\in \mathbb{R}^J$,  each unit $u=1, \dots,U$ computes its activation $h_{b,u}$ via different weights $\boldsymbol w_{b,u}\in \mathbb{R}^J$, i.e., $h_{b,u} = \boldsymbol w_{b,u}^T\boldsymbol x$. The linear responses of the units are concatenated, such that  $\boldsymbol h_b = [h_{b,1}, \dots, h_{b,U}]$, and transformed into probabilities via the softmax operation. Then, a Discrete sample $\boldsymbol \xi_b = [\xi_{b,1},\dots, \xi_{b,U}]$ is drawn; this constitutes an one-hot vector with a single non-zero entry at position $u'$, denoting the winner unit in the block. This winner unit, $u'$, passes its linear response to the next layer; the rest pass zero values.
		}
		\label{fig:block}
	\end{figure}

	In this context, to encode the winner unit in each of the $B$ LWTA blocks that constitute a stochastic LWTA layer, we introduce an appropriate set of discrete latent indicator vectors $\boldsymbol \xi \in \mathrm{one\_hot}(U)^B$. This vector comprises $B$ component subvectors, where each component entails \textit{exactly one non-zero value} at the \textit{index position} that corresponds to the \textit{winner unit} in each respective LWTA block. 
	
	Thus, the output $\boldsymbol y$ of a stochastic LWTA layer's $(b, u)$\textsuperscript{th} component $y_{b,u}$ is defined as:
	\begin{align}
		y_{b,u} = \xi_{b,u} \sum_{j=1}^J w_{j,b,u} \cdot x_j \in \mathbb{R}
		\label{eqn:y}
	\end{align}
	where $\xi_{b,u}$ denotes the $u$\textsuperscript{th} component of $\boldsymbol \xi_b$, and $\boldsymbol \xi_b \in \mathrm{one\_hot}(U)$ holds the $b$\textsuperscript{th} subvector of $\boldsymbol \xi$.

\begin{figure*}
		\begin{subfigure}[t]{.45\textwidth}
			\centering
			\resizebox{1.\linewidth}{!}{\def\layersep{3.5cm}
\def\inputsize{2}
\def\wtablocks{2}
\def\neuronsep{5}
\def\outputsize{2}
\def\wtasep{2.5}
\def\wtablocks{3}
\def\unitsperblock{2}
\def\prob{0.55}

\begin{tikzpicture}[-,draw, node distance=\layersep, label/.style args={#1#2}{%
    postaction={ decorate,
    decoration={ markings, mark=at position #1 with \node #2;}}}]
    \tikzstyle{every pin edge}=[<-,shorten <=1pt]
    \tikzstyle{neuron}=[circle,draw,minimum size=12pt,inner sep=0pt]
    \tikzstyle{wta} = [rounded corners,rectangle, draw, minimum_height=3cm, minimum_width=2cm]

    \tikzstyle{input neuron}=[neuron, fill=white!50];
    \tikzstyle{output neuron}=[neuron, fill=white];
    \tikzstyle{hidden neuron}=[neuron, fill=white!50];
    \tikzstyle{hidden neuron activated}=[neuron, fill=white!50, very thick];
    \tikzstyle{wta block} =[wta];
    \tikzstyle{annot} = [text width=10em, text centered]

    \node[input neuron, pin=left:$x_1$] (I-1) at (0,-\neuronsep+\wtasep) {};
    \node[input neuron, pin=left:$x_J$] (I-2) at (0,-2*\neuronsep+\wtasep) {};
		
     	\matrix (H-1) at (\layersep, -\wtasep-0.*\wtasep) [row sep=4mm, column sep=2mm, inner sep=3mm, box, matrix of nodes] 
      {
        		\node[hidden neuron activated](o1-1){}; \\
        		\node[hidden neuron](o1-2){}; \\
	  };

	  \matrix (H-2) at (\layersep, -3*\wtasep-0.*\wtasep) [row sep=4mm, column sep=2mm, inner sep=3mm, box, matrix of nodes] 
      {
        		\node[hidden neuron activated](o2-1){}; \\
        		\node[hidden neuron](o2-2){}; \\
	  };

	\matrix (H2-1) at (2*\layersep, -\wtasep-0.*\wtasep) [row sep=4mm, column sep=2mm, inner sep=3mm, box, matrix of nodes] 
      {
        		\node[hidden neuron](o21-1){}; \\
        		\node[hidden neuron activated](o21-2){}; \\
	  };

	  \matrix (H2-2) at (2*\layersep, -3*\wtasep-0.*\wtasep) [row sep=4mm, column sep=2mm, inner sep=3mm, box, matrix of nodes] 
      {
        		\node[hidden neuron](o22-1){}; \\
        		\node[hidden neuron activated](o22-2){}; \\
	  };

    \foreach \name / \y in {1,...,\outputsize}
    		\node[output neuron] (O-\name) at (3*\layersep,-\neuronsep*\y+\wtasep) {};

    \path (I-1) -- (I-2) node [black, font=\Huge, midway, sloped] {$\dots$};
 	\path[black!100 ] (I-1.east) edge  (o1-1);
    \path[black!100 ] (I-1.east) edge  (o1-2.west);
    
    \path[black!100 ] (I-1.east) edge (o2-1.west);
    \path[black!100 ] (I-1.east) edge (o2-2.west);

	\path[black!100 ] (I-2.east) edge (o1-1.west);
    \path[black!100 ] (I-2.east) edge (o1-2.west);
    
    \path[black!100] (I-2.east) edge  (o2-1.west);
    \path[black!100] (I-2.east) edge (o2-2.west);

    \path (H-1) -- (H-2) node [black, font=\Huge, midway, sloped] {$\dots$};
    \path (H2-1) -- (H2-2) node [black, font=\Huge, midway, sloped] {$\dots$};

    \path[color=black!100 ] (o1-1.east) edge (o21-1.west); 
    \path[color=black!100 ] (o1-1.east) edge (o21-2.west);

    \path[color=black!100] (o1-1.east) edge (o22-1.west);  
    \path[color=black!100] (o1-1.east) edge (o22-2.west); 
    
    \path[black!40] (o1-2.east) edge (o21-1.west);  
    \path[black!40] (o1-2.east) edge (o21-2.west);  
    
    \path[black!40] (o1-2.east) edge (o22-1.west);  
    \path[black!40] (o1-2.east) edge (o22-2.west);

    \path[color=black!100 ] (o2-1.east) edge (o21-1.west);  
    \path[color=black!100 ] (o2-1.east) edge (o21-2.west);
    
    \path[color=black!100 ] (o2-1.east) edge (o22-1.west);  
    \path[color=black!100 ] (o2-1.east) edge (o22-2.west);
    
    \path[black!40] (o2-2.east) edge (o21-1.west);  
    \path[black!40] (o2-2.east) edge (o21-2.west);
    
    \path[black!40] (o2-2.east) edge (o22-1.west);  
    \path[black!40] (o2-2.east) edge (o22-2.west);
    
    \path (O-1) -- (O-2) node [black, font=\Huge, midway, sloped] {$\dots$};

    \path[black!40] (o21-1.east) edge (O-1.west);  
    \path[black!40] (o21-1.east) edge (O-2.west);
    
    \path[black!100 ] (o21-2.east) edge (O-1.west);  
    \path[black!100 ] (o21-2.east) edge (O-2.west);

    \path[black!40] (o22-1.east) edge (O-1.west);  
    \path[black!40] (o22-1.east) edge (O-2.west);
    
    \path[black!100 ] (o22-2.east) edge (O-1.west);  
    \path[black!100 ] (o22-2.east) edge (O-2.west);

    \node[annot,above= -1mm of o1-1] (k) {\scriptsize $\xi$= 1};
    \node[annot,below= -1mm of o1-2] (k) {\scriptsize $\xi$= 0};
    
    \node[annot,above= -1mm of o22-1] (k) {\scriptsize $\xi$= 0};
    \node[annot,below= -1mm of o22-2] (k) {\scriptsize $\xi$= 1};
    \node[annot,above of=H-1, node distance=1.8cm] (hl) {LWTA layer};
    \node[annot,above left=-3mm and -1.7cm of H-1] (k) {\scriptsize $1$};
    
    \node[annot,above left=-3mm and -1.7cm of H2-1] (k) {\scriptsize $1$};
    
    \node[annot,below left=-3mm and -1.7cm of H-2] (k) {\scriptsize $B$};
    
    \node[annot,below left=-3mm and -1.7cm of H2-2] (k) {\scriptsize $B$};
    
    \node[annot,above of=H2-1, node distance=1.8cm] (hl2) {LWTA layer};
    \node[annot,left of=hl] {Input layer};
    \node[annot,right of=hl2] {Output layer};
\end{tikzpicture}
}
			\caption{}
			\label{synopsis:wta}
		\end{subfigure}
		\begin{subfigure}[t]{.55\textwidth}
			\centering
			\includegraphics[scale=0.5]{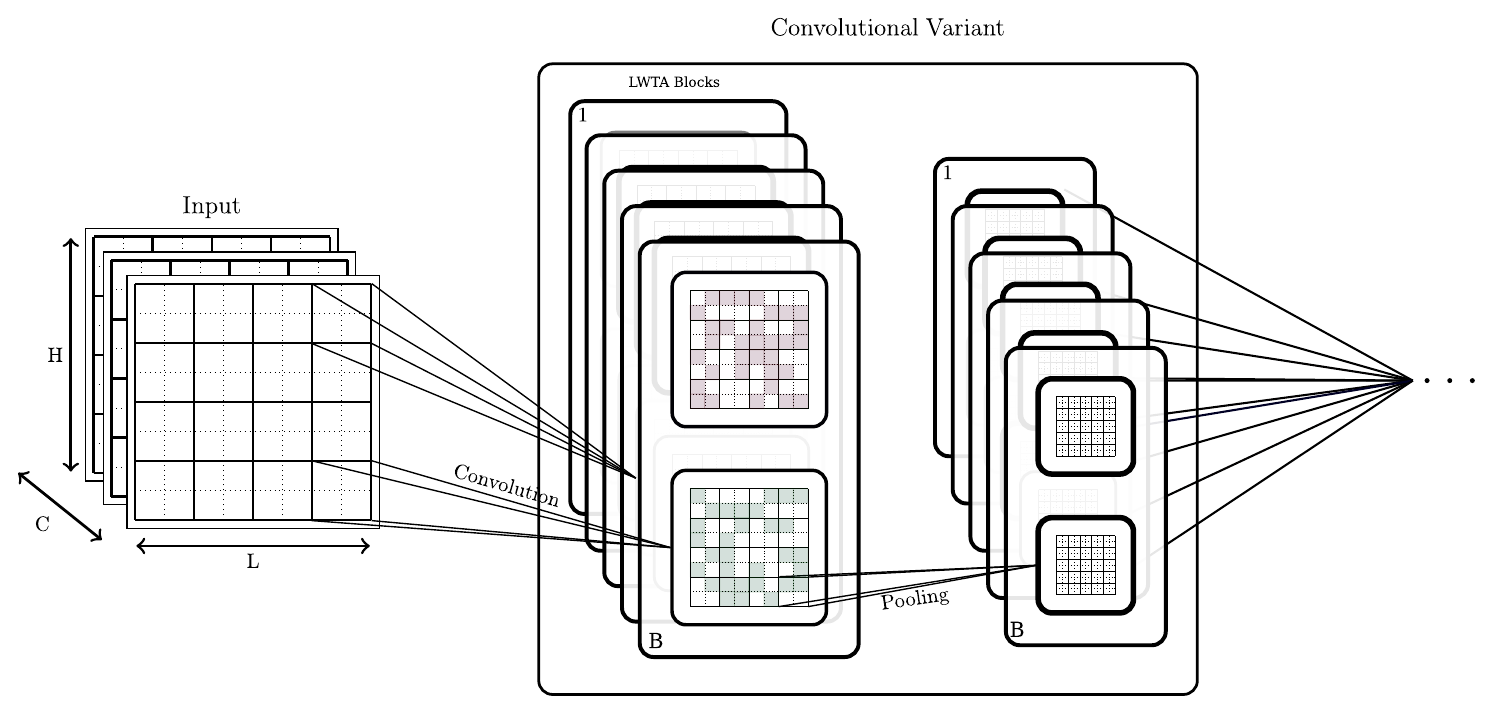}
			\caption{}
			\label{synopsis:fig:cnn_sb_lwta}
		\end{subfigure}
		\caption{ (a) A graphical representation of our competition-based modeling approach. Rectangles denote LWTA blocks, and circles the competing units therein. The winner units are denoted with bold contours ($\xi=1$). (b) The convolutional LWTA variant. Competition takes place \textit{position-wise} among the feature maps comprising a kernel. For each position, only the winner feature map contains a non-zero entry; for the rest feature maps in the kernel, the value at said position is zero.}
	\end{figure*}
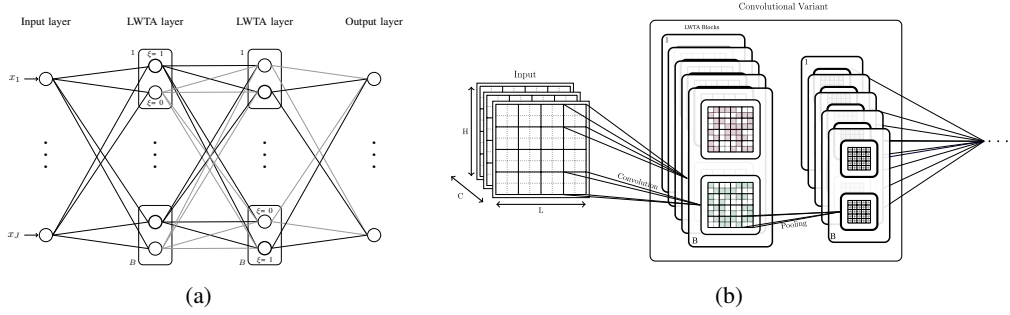

	We postulate that the latent winner indicator variables $\boldsymbol \xi_b, \forall b$ in Eq.\eqref{eqn:y} are drawn from a data-driven Categorical distribution with probabilities proportional to the intermediate linear computations that each unit performs. Therefore, the higher the linear response of a particular unit in a particular block, the higher its probability of it being the winner in said block; this yields:
		\begin{align}
			q(\boldsymbol \xi_b) = \mathrm{Categorical}\Big( \boldsymbol \xi_b \Big| \mathrm{softmax}\Big( \sum_{j=1}^J [w_{j,b,u}]_{u=1}^U \cdot  x_j \Big)
			\label{eqn:dense_q_xi}
		\end{align}
	where $[w_{j,b,u}]_{u=1}^U$ denotes the vector concatenation of the set $\{w_{j,b,u}\}_{u=1}^U$. A graphical illustration of the proposed stochastic LWTA block is depicted in Fig. \ref{fig:block}. Each stochastic LWTA layer comprises multiple such LWTA blocks, as illustrated in Fig. \ref{synopsis:wta}. Note that stochastically selecting the winner of each LWTA block in each layer introduces stochasticity to the network activations. Presented with the same input, different \textit{subnetworks} may be activated and a different \textit{subpath} is followed to the output, as a result of winner sampling.

	Further, to also accommodate in our study architectures based on the convolutional operation, we adopt the corresponding variant of the Stochastic LWTA activation in \cite{panousis2019nonparametric}.  We assume an input tensor $\boldsymbol{X} \in \mathbb{R}^{H\times L \times C}$ and define a set of kernels, each with weights  $\boldsymbol{W}_b\in \mathbb{R}^{h\times l \times C\times U}$, where $h, l, C, U$ are the kernel height, length, channels and competing feature maps, and $b=1, \dots, B$. Analogously to the grouping of linear units in dense layers, in this case, local competition is performed among feature maps on a \emph{position-wise} basis. Each kernel is treated as an LWTA block with competing feature maps; each layer comprises $B$ kernels. Specifically, each feature map $u=1, \dots, U$ in the $b$\textsuperscript{th} LWTA block of a convolutional LWTA layer computes:
	\begin{align}
	   \boldsymbol H_{b,u} = \boldsymbol W_{b,u} \star \boldsymbol X \in \mathbb{R}^{H \times L} 
    \end{align}
	Then, competition takes place among competing feature maps on a \textit{position-wise} basis. The competitive random sampling procedure reads:
	\begin{align}
    	q(\boldsymbol\xi_{b,h',l'}) = \mathrm{Categorical}\left( \boldsymbol\xi_{b,h',l'} \ \Big|\mathrm{softmax}\left(\left[ \boldsymbol H_{b,1,h',l'}, \dots, \boldsymbol H_{b,U, h',l'} \right]\right)\right), \ \forall h',l'
    	\label{eqn:conv_q_xi}
    \end{align}
	In each kernel, $b=1,\dots,B$, for each position, only the winner feature map contains a non-zero entry; all the rest feature maps contain zero values at this position. This yields sparse feature maps with mutually exclusive active positions. 
	
	Thus, at a given layer of the proposed convolutional variant, the output $\boldsymbol Y \in \mathbb{R}^{H \times L \times B \cdot U}$ is obtained via concatenation of the subtensors $\boldsymbol{Y}_{b,u}$ that read:
	\begin{align}
		\boldsymbol{Y}_{b,u} = \boldsymbol{\Xi}_{b,u} \Big( \boldsymbol{W}_{b, u}  \star \boldsymbol{X}\Big), \ \forall b,u
	\end{align}
	where $\boldsymbol{\Xi}_{b,u} = [\xi_{b,u,h',l'}]_{h',l'=1}^{H,L}$.
	\begin{figure}
		\centering
		\includegraphics[scale=.6]{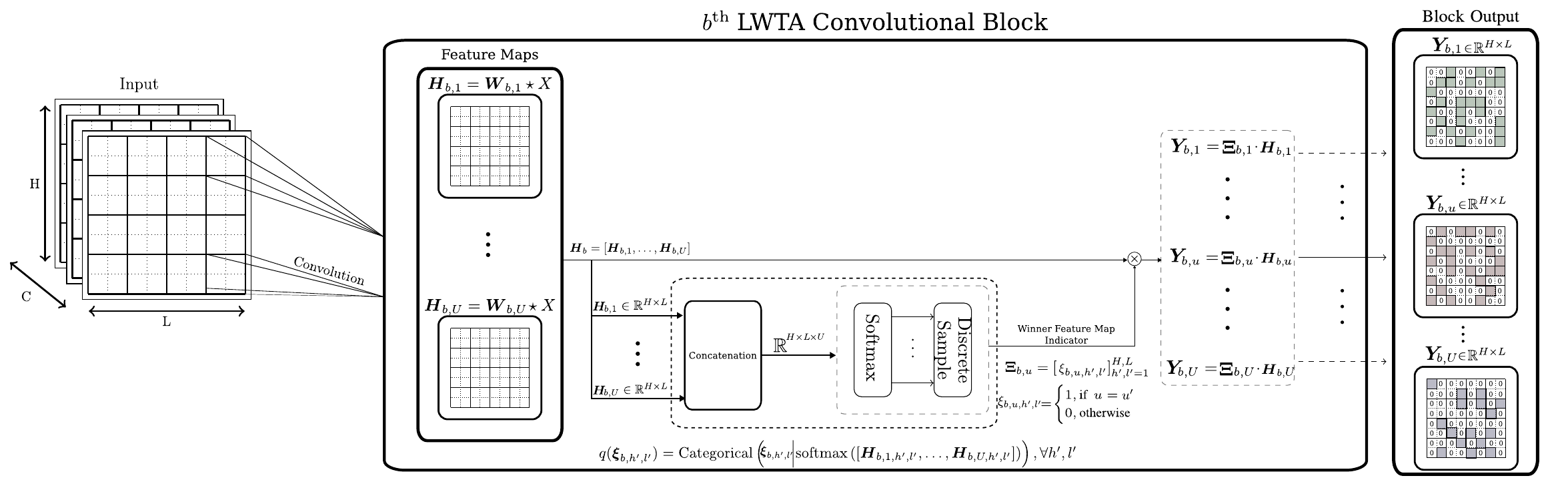}
		\caption{A detailed bisection of the $b$\textsuperscript{th} convolutional stochastic LWTA block. Presented with an input $\boldsymbol X \in \mathbb{R}^{H \times L \times C}$, competition now takes place among feature maps on a position-specific basis. Only the winner feature map contains a non-zero entry in a specific position. This leads to sparse feature maps, each comprising uniquely position-wise activated pixels.
		}
		\label{fig:block_conv}
	\end{figure}
	The corresponding illustration of the proposed stochastic convolutional LWTA block is depicted in Fig. \ref{fig:block_conv}. Convolutional Stochastic LWTA-based layers comprise multiple such blocks as shown in Fig. \ref{synopsis:fig:cnn_sb_lwta}.
	
	\subsection{Training}
	
	Since a network composed of such stochastic LWTA layers entails latent variables $\boldsymbol \xi$, we resort to a Bayesian treatment to perform effective parameter estimation. To this end, we turn to a stochastic gradient variational Bayes treatment \citep{kingma2014autoencoding} for scalability. The resulting objective takes the form of an evidence lower-bound (ELBO) as described next.

	Considering data $\mathcal{D} = \{X_i, Y_i\}_{i=1}^N$, we define the categorical cross-entropy between the data labels $Y_i$ and the class probabilities $f(X_i;\hat{\boldsymbol\xi})$, generated by the penultimate Softmax layer of a Stochastic LWTA-based network, as $\mathrm{CE}(Y_i, f(X_i;\hat{\boldsymbol\xi}))$. Here, $\hat{\boldsymbol\xi}$ denotes \textit{sample instances} of all the latent variables, $\boldsymbol\xi$, in all layers. We stress that the output class probabilities depend on the winner selection process in each layer, which is stochastic. 
	
	This way, the Evidence Lower Bound (ELBO) reads:
	\begin{align}
		\mathcal{L} = -\sum_{X_i, Y_i \in D} \mathrm{CE}(Y_i, f(X_i;\hat{\boldsymbol\xi})) - \mathrm{KL}[q(\boldsymbol\xi)||p(\boldsymbol\xi)])
		\label{eqn:elbo}
	\end{align} 
	For simplicity, and without loss of generality, we consider a symmetric Categorical distribution for the latent variable indicators $\boldsymbol \xi$; hence,  $p(\boldsymbol\xi_b) = \mathrm{Categorical}(1/U) \ \forall b$ for dense layers, and $p(\boldsymbol\xi_{b,h',l'}) = \mathrm{Categorical}(1/U) \ \forall b, h',l'$ for convolutional ones. In our work, we perform Monte-Carlo sampling using a single reparameterized sample for each of the corresponding latent variables. These are obtained via the reparameterization trick of the continuous relaxation of the Categorical distribution \citep{maddison, jang} as described next. We focus on the reparameterization trick for the dense case; the convolutional case is analogous.
	
	Let $\tilde{\xi}$ denote the probabilities of $q(\boldsymbol \xi)$ (Eqs. \eqref{eqn:dense_q_xi} and \eqref{eqn:conv_q_xi}). Then, the samples $\hat{\boldsymbol\xi}$ are expressed as:
	\begin{align}
	    \hat{\xi}_{b,u} = \mathrm{Softmax}((\log \tilde{\xi}_{b,u} + g_{b,u})/\tau), \ \forall b=1,\dots, B, \ u=1, \dots, U
	\end{align}
	where $g_{b,u} = -\log(-\log V_{b,u}), \ V_{b,u} \sim \mathrm{Uniform}(0,1)$, and $\tau \in (0, \infty)$ is a temperature factor, controlling how ``closely'' the continuous relaxation approximates the Categorical distribution. 

	On this basis, we can write the KL divergence term present in Eq. (\ref{eqn:elbo}) as:
	\begin{align}
		\begin{split}
		\mathrm{KL}[q(\boldsymbol\xi_{b})||p(\boldsymbol\xi_{b})] &= \mathbb{E}_{q(\boldsymbol\xi_b)}[\log q(\boldsymbol\xi_b) - \log p(\boldsymbol\xi_b)]\\
		&\approx \log q(\hat{\boldsymbol\xi}_b) - \log p(\hat{\boldsymbol\xi}_b), \ \forall b
		\end{split}
	\end{align}
	Hence, the final ELBO expression yields:
	\begin{align}
		\mathcal{L} = -\sum_{X_i, Y_i \in D} \mathrm{CE}(Y_i, f(X_i;\hat{\boldsymbol\xi})) - \sum_b \left( \log q(\hat{\boldsymbol\xi}_b) - \log p(\hat{\boldsymbol\xi}_b) \right)
	\end{align}
	
	\subsection{Prediction}
	
	At prediction time, we directly draw $L$ samples from the trained posteriors $q(\boldsymbol\xi)$ in order to determine the winning units in each block of the network. As mentioned before, each time we sample for the same input, a different \textit{subpath} is followed.  This is a key aspect that \textit{stochastically} alters the information flow in the network and \textit{obstructs} an adversary from attacking the model. 
	
	The sampling process results in a set of $L$ output logits of the network, which we can average to obtain the final prediction:
	\begin{align}
	    f(X_i; \hat{\boldsymbol\xi}) = \frac{1}{L}\sum_{l=1}^L f(X_i; \hat{\boldsymbol\xi}^l)
	\end{align}
    where $\hat{\boldsymbol\xi}^l$ denotes a sample drawn from  $q(\boldsymbol\xi)$. 
	\section{Experimental Results}
	
	We investigate the potency of LWTA-based networks against adversarial attacks under an Adversarial Training regime; we employ a PGD adversary \citep{madry2017towards}. To this end, we use the well-known WideResNet-34 \citep{wideresnet} architecture, considering three different widen factors: 1, 5, and 10; we focus on the CIFAR-10 dataset and adopt experimental settings similar to \cite{wu2021wider}. We use a batch-size of 128 and an initial learning rate of 0.1; we halve the learning rate at every epoch after the $75^{th}$ epoch. We use a single sample $L=1$ for prediction. All experiments were performed using a single NVIDIA Quadro P6000.
	
	For evaluating the robustness of the proposed structure, we initially consider the conventional PGD attack with 20 steps, step size $0.007$ and $\epsilon = 8/255$. In Table \ref{tab:trades_width_1}, we compare the robustness of LWTA-based WideResNet networks against the baseline results of \cite{wu2021wider}. As we observe, our Stochastic LWTA-based networks yield significant improvements in robustness under a traditional PGD attack; they retain extremely high natural accuracy (up to $\approx 13\%$ better), while exhibiting a staggering, up to $\approx 32.6\%$, difference in robust accuracy compared to the \textit{exact same architectures} employing the conventional ReLU-based nonlinearities and trained in the \textit{exact same fashion}.
	\begin{table}
		\caption{Natural and Robust accuracy under a conventional PGD attack with 20 steps and $0.007$ step-size using WideResNet-34 models with different widen factors. We use the same PGD-based Adversarial Training scheme for all models \citep{madry2017towards}.}
		\label{tab:trades_width_1}
		\renewcommand{\arraystretch}{1.1}
		\centering
		\resizebox{0.7\textwidth}{!}{
			\begin{tabular}{ccc|cc}
				\multicolumn{5}{c}{Adversarial Training-PGD}\\
				\hline
				& \multicolumn{2}{c}{Natural Accuracy ($\%$)} & \multicolumn{2}{c}{Robust Accuracy ($\%$)}\\
				\cline{2-3}\cline{4-5}
				Widen Factor & Baseline & Stochastic LWTA & Baseline & Stochastic LWTA \\\hline
				1  & 74.04 & \textbf{87.0} & 49.24 & \textbf{81.87} \\
				5  & 83.95 & \textbf{ 91.88} & 54.36 & \textbf{83.4} \\
				10 & 85.41 & \textbf{92.26} & 55.78 &  \textbf{84.3} \\\hline
			\end{tabular}
		}
	\end{table}

	\begin{table}[h!]
		\caption{Robust Accuracy $(\%)$ comparison under the AutoAttack framework. $\dagger$ denotes models that are trained with additional unlabeled data. The AutoAttack performance corresponds to the final robust accuracy after employing all the attacks in AA. Results directly from the AA leaderboard. }
		\label{tab:sota}
		\centering
		\resizebox{0.65\textwidth}{!}{%
			\renewcommand{\arraystretch}{1.1}
			\begin{tabular}{c|c}
				\hline
				Method & AutoAttack\\\hline
				TRADES\citep{zhang2019limitations} & 53.08\\
				Early-Stop \citep{rice2020} & 53.42\\
				FAT \citep{zhang2020} & 53.51\\
				HE \citep{pang2020} & 53.74\\
				WAR \citep{wu2021wider} & 54.73\\ \hhline{=|=}
				Pre-training \citep{hendrycks2019}$\dagger$ & 54.92\\
				MART \citep{wang2020}$\dagger$ & 56.29\\
				HYDRA \citep{Sehwag2020}$\dagger$ & 57.14\\
				RST \citep{Carmon2019}$\dagger$ & 59.53\\
				\cite{gowal2021uncovering}$\dagger$ & 65.88\\
				WAR \citep{wu2021wider}$\dagger$ & 61.84\\\hline
				Ours (Stochastic-LWTA/PGD/WideResNet-34-1) & \textbf{74.71}\\
				Ours (Stochastic-LWTA/PGD/WideResNet-34-5) & \textbf{81.22}\\
				Ours (Stochastic-LWTA/PGD/WideResNet-34-10) & \textbf{82.60}\\\hline
			\end{tabular}
		}
	\end{table}
	
	Further, and to ensure that our approach does not cause the well-known obfuscated gradient problem \citep{AthalyeC018}, we resort to stronger parameter-free attacks using the newly introduced AutoAttack (AA) framework \citep{croce2020reliable}. AA comprises an ensemble of four powerful white-box and black-box attacks, e.g., the commonly employed A-PGD attack; this is a step-free variant of the standard PGD attack \citep{madry2017towards}, which avoids the complexity and ambiguity of step-size selection. In addition, for the entailed $L_\infty$ attack, we use the common $\epsilon = 8/255$ value. Thus, in Table \ref{tab:sota}, we compare the LWTA-based networks to several recent state-of-the-art approaches evaluated on AA\footnote{\url{https://github.com/fra31/auto-attack}}. The reported accuracies correspond to the final  reported robust accuracy of the methods after sequentially performing all the considered AA attacks.  Once again, we observe that the proposed networks yield state-of-the-art robustness against all SOTA methods, with an improvement of $\approx 16.72\%$, even when compared with methods that employ substantial data augmentation to increase robustness, e.g. \cite{gowal2021uncovering}. These results vouch for the potency of Stochastic LWTA networks in adversarial settings. 
	
	Finally, since our considered networks consist of stochastic components, i.e. the competitive random sampling procedure to determine the winner in each LWTA block, the output of the classifier might change at each iteration; this obstructs the attacker from successfully altering the final decision. To counter such randomness in the involved computations, \cite{croce2020reliable} combine the APGD attack with an averaging procedure of 20 computations of the gradient at the same point. This technique is known as Expectation over Transformation (EoT) \citep{AthalyeC018}. Thus, we use AA combined with EoT for further performance evaluation of the proposed LWTA-based networks. The corresponding results are presented in Table \ref{tab:eot}. As we observe, all of the considered networks retain state-of-the-art robustness against the powerful AA \& EoT attacks. This conspicuously supports the usefulness of Stochastic LWTA activations towards adversarial robustness.

	\begin{table}[h!]
		\caption{Robustness against AA combined with 20 iterations of EoT. APGD-DLR corresponds to the APGD attack, using a different loss, i.e., the Difference of Logits Ratio \citep{croce2020reliable}.}
		\label{tab:eot}
		\renewcommand{\arraystretch}{1.1}
		\centering
		\begin{tabular}{ccccc}
			\hline
			Widen Factor & Nat. Acc. & APGD & APGD-DLR\\\hline
			1  & 87.00 & 79.67 & 76.15  \\
			5  & 91.88 & 81.67 & 77.65 \\
			10  & 92.26 & 82.55 & 79.00 \\\hline
		\end{tabular}
	\end{table}

	\section{Conclusions}
	In this work, we explored the potency of Stochastic LWTA-based networks against powerful white-box and black-box attacks. The experimental results vouch for the efficacy of the arising networks, yielding state-of-the-art robustness in all the experimental settings. We obtained an immense improvement in robustness compared to the second best performing alternative, which notably relies on substantial data augmentation. A potentially key principle towards adversarial robustness may be the stochastic alteration of the information flow in Stochastic LWTA-based networks; this, arises from the considered data-driven winner selection mechanism in each LWTA block. Different subpaths, stochastically emerging even for the same input, essentially obstruct the adversary from successfully attacking the model. Further evaluation against stochasticity countermeasures, i.e., Expectation over Transformation (EoT), further validate our findings, as the induced decrease in the final robustness was negligible.

	\section*{Acknowledgements}
	
	This work has received funding from the European Union's Horizon 2020 research and innovation program
	under grant agreement No 872139, project aiD.
	
	\interlinepenalty=10000
	
	\bibliography{lwta_nips}
	
\end{document}